\documentclass[11pt,letterpaper]{article}
\usepackage{naaclhlt2013}
\usepackage{times}
\usepackage{latexsym}
\usepackage{graphicx}
\usepackage{lscape}
\usepackage{enumitem}
\usepackage{url}
\usepackage{comment}

\usepackage{covington}
\usepackage{color}

\title{SemEval-2013 Task 2: Sentiment Analysis in Twitter}

\author{
{\bf Preslav Nakov}\\QCRI, Qatar Foundation\\{\small {\tt pnakov@qf.org.qa}}\\
{\bf Zornitsa Kozareva}\\USC Information Sciences Institute\\{\small {\tt kozareva@isi.edu}}\\
{\bf Alan Ritter}\\University of Washington\\{\small {\tt aritter@cs.washington.edu}}\\
\And
{\bf Sara Rosenthal }\\Columbia University\\{\small {\tt sara@cs.columbia.edu}}\\
{\bf Veselin Stoyanov}\\JHU HLTCOE\\{\small {\tt ves@cs.jhu.edu}}\\
{\bf Theresa Wilson}\\JHU HLTCOE\\{\small {\tt taw@jhu.edu}}\\
}

\date{}

\begin{document}
\maketitle

\begin{abstract}

In recent years, sentiment analysis in social media has attracted a lot of research interest and has been used for a number of applications. Unfortunately, research has been hindered by the lack of suitable datasets, complicating the comparison between approaches. To address this issue, we have proposed \emph{SemEval-2013 Task 2: Sentiment Analysis in Twitter},
which included two subtasks: A, an expression-level subtask, and B, a message-level subtask. We used crowdsourcing on Amazon Mechanical Turk to label a large Twitter training dataset along with additional test sets of Twitter and SMS messages for both subtasks. All datasets used in the evaluation are released to the research community.
The task attracted significant interest and a total of 149 submissions from 44 teams. The best-performing team achieved an F1 of 88.9\% and 69\% for subtasks A and B, respectively.

\end{abstract}

\section{Introduction}
\label{sec-intro}

In the past decade, new forms of communication, such as microblogging and text messaging have emerged and become ubiquitous.  Twitter messages (tweets) and cell phone messages (SMS) are often used to share opinions and sentiments about the surrounding world, and the availability of social content generated on sites such as Twitter creates new opportunities to automatically study public opinion.

Working with these informal text genres presents new challenges for natural language processing
beyond those encountered when working with more traditional text genres such as newswire.

Tweets and SMS messages are short in length: a sentence or a headline rather than a document.  The language they use is very informal, with creative spelling and punctuation, misspellings, slang, new words, URLs, and genre-specific terminology and abbreviations, e.g., RT for re-tweet and \#hashtags.\footnote{Hashtags are a type of tagging for Twitter messages.}  How to handle such challenges so as to automatically mine and understand the opinions and sentiments that people are communicating has only very recently been the subject of research \cite{Jansen09,Barbosa10,Bifet11,Davidov10,oconnor10,Pak10,Tumasjan10,Kouloumpis11}.


Another aspect of social media data, such as Twitter messages, is that they include rich structured information about the individuals involved in the communication. For example, Twitter maintains information about who follows whom. Re-tweets (re-shares of a tweet) and tags inside of tweets provide discourse information. Modeling such structured information is important because it provides means for empirically studying social interactions where opinion is conveyed, e.g., we can study the properties of persuasive language or those associated with influential users.

 \begin{table*}[t]
\small
\begin{tabular}{|l|p{14cm}|}
\hline
\bf Twitter & RT @tash\_jade: That's really sad, Charlie RT ``Until tonight I never realised how fucked up I was" - Charlie Sheen \#sheenroast  \\
\hline
\bf SMS & Glad to hear you are coping fine in uni... So, wat interview did you go to? How did it go?\\
 \hline
 \end{tabular}
\caption{Examples of sentences from each corpus that contain subjective phrases.}
\label{T:Sentences}
\end{table*}

Several corpora with detailed opinion and sentiment annotation have been made freely available, e.g., the MPQA corpus \cite{Wiebe05} of newswire text.
These corpora have proved very valuable as resources for learning about the language of sentiment in general,
but they did not focus on social media.

While some Twitter sentiment datasets have already been created,
they were either small and proprietary, such as the i-sieve corpus \cite{Kouloumpis11}, or they were created only for Spanish
like the TASS corpus\footnote{http://www.daedalus.es/TASS/corpus.php} \cite{TASS},
or they relied on noisy labels obtained from emoticons and hashtags.
They further focused on message-level sentiment,
and no Twitter or SMS corpus with expression-level sentiment annotations has been made available so far.

Thus, the primary goal of our SemEval-2013 task 2 has been to promote research
that will lead to a better understanding of how sentiment is conveyed in Tweets and SMS messages.
Toward that goal, we created the SemEval Tweet corpus,
which contains Tweets (for both training and testing) and SMS messages (for testing only) with sentiment expressions
annotated with contextual phrase-level polarity as well as an overall message-level polarity.
We used this corpus as a testbed for the system evaluation at SemEval-2013 Task 2.

In the remainder of this paper, we first describe the task, the dataset creation process, and the evaluation methodology.
We then summarize the characteristics of the approaches taken by the participating systems and we discuss their scores.

\section{Task Description}
\label{sec-decription}

We had two subtasks: an expression-level subtask and a message-level subtask.
Participants could choose to participate in either or both subtasks.
Below we provide short descriptions of the objectives of these two subtasks.

\begin{description}
  \item[Subtask A: Contextual Polarity Disambiguation] Given a message containing a marked instance of a word or a phrase, determine whether that instance is positive, negative or neutral in that context. The boundaries for the marked instance were provided:
      this was a classification task, not an entity recognition task.
  \item [Subtask B: Message Polarity Classification]
  Given a message, decide whether it is of positive, negative, or neutral sentiment.
    For messages conveying both a positive and a negative sentiment, whichever is the stronger one was to be chosen.
\end{description}

Each participating team was allowed to submit results for two different systems per subtask: one constrained, and one unconstrained.
A constrained system could only use the provided data for training, but it could also use other resources such as lexicons obtained elsewhere.
An unconstrained system could use any additional data as part of the training process;
this could be done in a supervised, semi-supervised, or unsupervised fashion.

  \begin{figure*}
\framebox{\parbox[t]{16.2cm}{\small\textbf{Instructions:} Subjective words are ones which convey an opinion. Given a sentence, identify whether it is objective, positive, negative, or neutral. Then, identify each subjective word or phrase in the context of the sentence and mark the position of its start and end in the text boxes below. The number above each word indicates its position. The word/phrase will be generated in the adjacent textbox so that you can confirm that you chose the correct range. Choose the polarity of the word or phrase by selecting one of the radio buttons: positive, negative, or neutral. If a sentence is not subjective please select the checkbox indicating that "There are no subjective words/phrases". Please read the examples and invalid responses before beginning if this is your first time answering this hit.}}
\framebox{\parbox[t]{16.2cm}{\includegraphics[scale=.37]{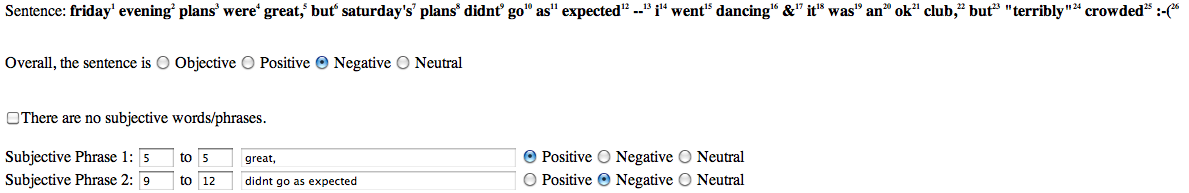}}}
\caption{Instructions provided to workers on Mechanical Turk followed by a screenshot.}
\label{F:instructions}
\end{figure*}

Note that constrained/unconstrained refers to the data used to train a classifier. For example, if other data (excluding the test data) was used to develop a sentiment lexicon, and the lexicon was used to generate features, the system would still be constrained. However, if other data (excluding the test data) was used to develop a sentiment lexicon, and this lexicon was used to automatically label additional Tweet/SMS messages and then used with the original data to train the classifier, then such a system would be unconstrained.

\section{Dataset Creation}
\label{sec-dataset}

 \begin{table*}[t]
\small
\begin{center}
\begin{tabular}{|l|r|r|r|r|r|r|}
\hline
 & \multicolumn{2}{c|}{\bf Average \# of} & \multicolumn{3}{c|}{\bf Total Phrase Count} & \bf Vocabulary \\
\bf Corpus & \bf Words & \bf Characters & \bf Positive & \bf Negative & \bf Neutral & \multicolumn{1}{c|}{\bf Size}\\
\hline
Twitter - Training &25.4  &        120.0        &       5,895  &  3,131 &  471 &  20,012     \\
Twitter - Dev & 25.5 & 120.0 & 648 & 430 & 57 & 4,426 \\
Twitter - Test  & 25.4 & 121.2 & 2,734 & 1,541 & 160 & 11,736 \\
SMS - Test & 24.5 & 95.6 & 1,071 & 1,104 & 159 & 3,562 \\
\hline
\end{tabular}
\caption{Statistics for Subtask A.}
\label{T:CorpusStatsA}
\end{center}
\end{table*}

In the following sections we describe the collection and annotation of the Twitter and SMS datasets.

 \subsection{Data Collection}

Twitter is the most common micro-blogging site on the Web,
and we used it to gather tweets that express sentiment about popular topics.
We first extracted named entities using a Twitter-tuned NER system \cite{ner}
from millions of tweets, which we collected over a one-year period spanning from January 2012 to January 2013;
we used the public streaming Twitter API to download tweets.

We then identified popular topics as those named entities that are frequently mentioned in association with a specific date \cite{ritter12}.
Given this set of automatically identified topics, we gathered tweets from the same time period which mentioned the named entities.
The testing messages had different topics from training and spanned later periods.

To identify messages that express sentiment towards these topics, we filtered the tweets using SentiWordNet \cite{swn}.
We removed messages that contained no sentiment-bearing words, keeping only those with at least one word with positive or negative sentiment score
that is greater than 0.3 in SentiWordNet for at least one sense of the words.
Without filtering, we found class imbalance to be too high.\footnote{Filtering based on an existing lexicon does bias the dataset to some degree; however, note that the text still contains sentiment expressions outside those in the lexicon.}

 \begin{table}[h]
\small
\begin{center}
\begin{tabular}{|l |r |r | r |}
\hline
\bf Corpus & \bf Positive & \bf Negative & \bf Objective\\
& & & \bf / Neutral\\
\hline
Twitter - Training  & 3,662 & 1,466 &  4,600  \\
Twitter - Dev  & 575 & 340 & 739 \\
Twitter - Test  & 1,573 & 601 & 1,640   \\
SMS - Test  & 492 & 394 & 1,208  \\
\hline
\end{tabular}
\caption{Statistics for Subtask B.}
\label{T:CorpusStatsB}
\end{center}
\end{table}

Twitter messages are rich in social media features, including out-of-vocabulary (OOV) words, emoticons, and acronyms;
see Table~\ref{T:Sentences}.
A large portion of the OOV words are hashtags (e.g., \texttt{\#sheenroast})
and mentions (e.g., \texttt{@tash\_jade}).

We annotated the same Twitter messages with annotations for subtask A and subtask B.
However, the final training and testing datasets overlap only partially between the two subtasks
since we had to throw away messages with low inter-annotator agreement, and this differed between the subtasks.
For testing, we also annotated SMS messages, taken from the NUS SMS corpus\footnote{\texttt{http://wing.comp.nus.edu.sg/SMSCorpus/}} \cite{SMScorpus}.
Tables~\ref{T:CorpusStatsA} and ~\ref{T:CorpusStatsB} show statistics about the corpora we created for subtasks A and B.

 \subsection{Annotation Guidelines}

The instructions provided to the annotators, along with an example, are shown in Figure~\ref{F:instructions}.
We provided several additional examples to the annotators, shown in Table~\ref{T:annotation-examples}.

 \begin{table}[t]
\small
\begin{center}
\begin{tabular}{| l | l | l | l | l |}
\hline
& \multicolumn{3}{|c|}{\bf A} & \multicolumn{1}{|c|}{\bf B} \\
\hline
& \bf Lower & \bf Avg. & \bf Upper & \bf Avg. \\
\hline
Twitter - Train & 64.7 & 82.4 & 90.8 & 82.7\\
Twitter - Dev & 51.2 & 74.7 & 87.8 & 78.4\\
Twitter - Test & 68.8 & 83.6 & 90.9 & 76.9\\
SMS - Test & 66.5 & 88.5 & 81.2 & 77.6\\
\hline
\end{tabular}
\caption{Bounds for datasets in subtasks A and B.}
\label{T:bounds}
\end{center}
\end{table}

 \begin{table*}[t]
 \centering
\small
\begin{tabular}{|p{16cm}|}
\hline
 Authorities are \textit{\color{red}only too aware} that Kashgar is 4,000 kilometres (2,500 miles) from Beijing but \textit{only} a tenth of the distance from the  Pakistani border, and are \textit{\color{red}desperate} to \textit{\color{red} ensure instability or militancy} does not leak over the frontiers. \\
 \hline
Taiwan-made products \textit{\color{green} stood a good chance} of becoming \textit{\color{green} even more competitive thanks to} wider access to overseas markets and lower costs for material imports, he said. \\
\hline
"March \textit{\color{blue}appears} to be a \textit{\color{blue}more reasonable} estimate while earlier admission \textit{\color{blue}cannot be entirely ruled out}," according to Chen, also Taiwan's chief WTO negotiator. \\
\hline
friday evening plans were great, but saturday's plans \textit{\color{red}didnt go as expected} -- i went dancing \& it was an \textit{\color{blue}ok} club, but \textit{\color{red}terribly crowded :-(} \\
\hline
WHY THE \textit{\color{red}HELL} DO YOU GUYS ALL HAVE MRS. KENNEDY! SHES A FUCKING DOUCHE \\
\hline
AT\&T was \textit{\color{blue}okay} but whenever they do something \textit{\color{green}nice} in the name of customer service it seems like a favor, while T-Mobile makes that a \textit{\color{green}normal everyday thin} \\
\hline
obama should be \textit{\color{red}impeached} on \textit{\color{red}TREASON} charges. Our Nuclear arsenal was TOP Secret. Till HE told our enemies what we had. \textit{\color{red}\#Coward \#Traitor} \\
\hline
My graduation speech: "I'd like to \textit{\color{green}thanks} Google, Wikipedia and my computer! \textit{\color{green}:D} \#iThingteens \\
\hline
\end{tabular}
\caption{List of example sentences with annotations that were provided to the annotators. All subjective phrases are italicized. Positive phrases are in green, negative phrases are in red, and neutral phrases are in blue.}
\label{T:annotation-examples}
\end{table*}

In addition, we filtered spammers by considering the following kinds of annotations invalid:

\begin{itemize}[noitemsep]
\item containing overlapping subjective phrases;
\item subjective but without a subjective phrase;
\item marking every single word as subjective;
\item not having the overall sentiment marked.
\end{itemize}

\subsection{Annotation Process}

Our datasets were annotated for sentiment on Mechanical Turk. Each sentence was annotated by five Mechanical Turk workers (Turkers). In order to qualify for the hits, the Turker had to have an approval rate greater than 95\% and have completed 50 approved hits. Each Turker was paid three cents per hit. The Turker had to mark all the subjective words/phrases in the sentence by indicating their start and end positions and say whether each subjective word/phrase was positive, negative, or neutral (subtask A). They also had to indicate the overall polarity of the sentence (subtask B).

Figure~\ref{F:instructions} shows the instructions and an example provided to the Turkers.
The first five rows of Table~\ref{T:Annotation} show an example of the subjective words/phrases
marked by each of the workers.

For subtask A, we combined the annotations of each of the workers using intersection as indicated in the last row of Table~\ref{T:Annotation}. A word had to appear in 2/3 of the annotations in order to be considered subjective.  Similarly, a word had to be labeled with a particular polarity (positive, negative, or neutral) 2/3 of the time in order to receive that label.

We also experimented with combining annotations by computing the union of the sentences, and taking the sentence of the worker who annotated the most hits, but we found that these methods were not as accurate. Table~\ref{T:bounds} shows the lower, average, and upper bounds for all the hits by computing the bounds for each hit and averaging them together. This gives a good indication about how well we can expect the systems to perform. For example, even if we used the best annotator each time, it would still not be possible to get perfect accuracy.

For subtask B, the polarity of the entire sentence was determined based on the majority of the labels. If there was a tie, the sentence was discarded. In order to reduce the number of sentences lost, we combined the objective and the neutral labels, which Turkers tended to mix up. Table~\ref{T:bounds} shows the average bound for subtask B by computing the bounds for each hit and averaging them together. Since the polarity is chosen based on the majority, the upper bound is 100\%.

 \begin{table*}[t]
\centering
\small
\begin{tabular}{|l|l|l|}
\hline
Worker 1 & \textbf{\textit{I would love}} to watch Vampire Diaries \textbf{\textit{:)}} and some Heroes! \textbf{\textit{Great combination}} & 9/13 \\
Worker 2 & I would love to watch Vampire Diaries :) and some \textbf{\textit{Heroes!}} \textbf{\textit{Great}} combination & 11/13 \\
Worker 3 & I \textbf{\textit{would love}} to watch Vampire Diaries \textbf{\textit{:)}} and some Heroes! \textbf{\textit{Great}} combination & 10/13 \\
Worker 4 & I would \textbf{\textit{love}} to watch Vampire Diaries :) and some Heroes! \textbf{\textit{Great}} combination & 13/13 \\
Worker 5 & I would love to watch Vampire Diaries :) and some Heroes! \textbf{\textit{Great}} combination & 11/13\\
\hline
Intersection & I would \textbf{\textit{love}} to watch Vampire Diaries :) and some Heroes! \textbf{\textit{Great}} combination & \\
\hline
\end{tabular}
\caption{Example of a sentence annotated for subjectivity on Mechanical Turk. Words and phrases that were marked as subjective are italicized and highlighted in bold. The first five rows are annotations provided by Turkers, and the final row shows their intersection. The final column shows the accuracy for each annotation compared to the intersection.}
\label{T:Annotation}
\end{table*}

\section{Scoring}
\label{sec-scoring}

For both subtasks, the participating systems were required to perform a three-way classification -- a particular marked phrase (for subtask A) or an entire message (for subtask B) was to be classified as {\em positive}, {\em negative}, or {\em objective}.
For each system, we computed a score for predicting positive/negative phrases/messages vs. the other two classes.

For instance, to compute positive precision, $P_{pos}$, we find the number of phrases/messages that a system correctly predicted to be positive,
and we divide that number by the total number of messages it predicted to be positive.
To compute recall, for the positive class, $R_{pos}$, we find the number of messages correctly predicted to be positive and we divide that number by the total number of positive messages in the gold standard.

We then calculate F-score for the positive labels, the harmonic average of precision and recall as follows $F_{pos}= 2 \frac{P_{pos} R_{pos}}{P_{pos}+R_{pos}}$.
We carry out a similar computation to calculate $F_{neg}$, which is F1 for negative messages.

The overall score for each system run is then given by the average of the F1-scores for the positive and negative classes: $F=(F_{pos}+F_{neg})/2$.

Note that ignoring $F_{neutral}$ does not reduce the task to predicting positive vs. negative labels only (even though some participants have chosen to do so)
since the gold standard still contains neutral labels which are to be predicted:
$F_{pos}$ and $F_{neg}$ would suffer if these examples are labeled as positive and/or negative instead of neutral.

We provided participants with a scorer. In addition to outputting the overall F-score, it produced a confusion matrix for the three prediction classes ({\em positive}, {\em negative}, and {\em objective}), and it also validated the data submission format.

\begin{table}[hbt]
    \centering
    \begin{small}
    \begin{tabular}{@{ }lc@{ }c@{ }cc@{ }}
    \bf Run & \bf Const- & \bf Unconst- & \bf Use & \bf Super-\\
    & \bf rained & \bf rained & \bf Neut.? & \bf vised?\\
    \hline
    NRC-Canada & 88.93 &  & yes & yes\\
    AVAYA & 86.98 & 87.38$_{(1)}$ & yes & yes\\
    BOUNCE & 86.79 &  & yes & yes\\
    LVIC-LIMSI & 85.70 &  & yes & yes\\
    FBM & 85.50 &  & yes & semi\\
    GU-MLT-LT & 85.19 &  & yes & yes\\
    $^\diamond$UNITOR & 84.60 &  & yes & yes\\
    USNA & 81.31 &  & yes & yes\\
    Serendio & 80.04 &  & yes & yes\\
    $^\diamond$ECNUCS & 79.48 & 80.15$_{(2)}$ & yes & yes\\
    TJP & 78.16 &  & yes & yes\\
    $^\circ$columbia-nlp & 74.94 &  & yes & yes\\
    teragram &  & 74.89$_{(3)}$ & yes & yes\\
    sielers & 74.41 &  & yes & yes\\
    KLUE & 73.74 &  & yes & yes\\
    OPTWIMA & 69.17 & 36.91$_{(6)}$ & yes & yes\\
    swatcs & 67.19 & 63.86$_{(5)}$ & no & yes\\
    Kea & 63.94 &  & yes & yes\\
    senti.ue-en & 62.79 & 71.38$_{(4)}$ & yes & yes\\
    uottawa & 60.20 &  & yes & yes\\
    IITB & 54.80 &  & yes & yes\\
    SenselyticTeam & 53.88 &  & yes & yes\\
    SU-sentilab &  & 34.73$_{(7)}$ & no & yes\\
        \hline
    Majority Baseline & \multicolumn{2}{c}{38.10} & N/A & N/A \\
    \hline
    \end{tabular}
    \end{small}
    \caption{\label{tab:TaskATwitter}Results for subtask A on the Twitter dataset.
             The $^\circ$ marks a team that includes a task coorganizer,
             and the $^\diamond$ indicates a system submitted as constrained but which used additional Tweets or additional sentiment-annotated text
             to collect statistics that were then used as a feature.}
  \end{table}

\begin{table}[hbt]
    \centering
    \begin{small}
    \begin{tabular}{@{ }lc@{ }c@{ }cc@{ }}
    \bf Run & \bf Const- & \bf Unconst- & \bf Use & \bf Super-\\
    & \bf rained & \bf rained & \bf Neut.? & \bf vised?\\
    \hline
    GU-MLT-LT & 88.37 &  & yes & yes\\
    NRC-Canada & 88.00 &  & yes & yes\\
    $^\star$AVAYA & 83.94 & 85.79$_{(1)}$ & yes & yes\\
    $^\diamond$UNITOR & 82.49 &  & yes & yes\\
    TJP & 81.23 &  & yes & yes\\
    LVIC-LIMSI & 80.16 &  & yes & yes\\
    USNA & 79.82 &  & yes & yes\\
    $^\diamond$ECNUCS & 76.69 & 77.34$_{(2)}$ & yes & yes\\
    sielers & 73.48 &  & yes & yes\\
    FBM & 72.95 &  & no & semi\\
    teragram & 72.83 & 72.83$_{(4)}$ & yes & yes\\
    KLUE & 70.54 &  & yes & yes\\
    $^\circ$columbia-nlp & 70.30 &  & yes & yes\\
    senti.ue-en & 66.09 & 74.13$_{(3)}$ & yes & yes\\
    swatcs & 66.00 & 67.68$_{(5)}$ & no & yes\\
    Kea & 63.27 &  & yes & yes\\
    uottawa & 55.89 &  & yes & yes\\
    SU-sentilab &  & 55.38$_{(6)}$ & no & yes\\
    SenselyticTeam & 51.13 &  & yes & yes\\
    OPTWIMA & 37.32 & 36.38$_{(7)}$ & yes & yes\\
   \hline
    Majority Baseline & \multicolumn{2}{c}{31.50}  & N/A & N/A \\
    \hline
        \end{tabular}
    \end{small}
    \caption{\label{tab:TaskASMS}Results for subtask A on the SMS dataset. The $^\star$ indicates a late submission,
                the $^\circ$ marks a team that includes a task co-organizer,
             and the $^\diamond$ indicates a system submitted as constrained but which used additional Tweets or additional sentiment-annotated text
             to collect statistics that were then used as a feature.}
  \end{table}

\section{Participants and Results}
\label{sec-results}

The results for subtask A are shown in Tables \ref{tab:TaskATwitter} and \ref{tab:TaskASMS} for Twitter and for SMS messages, respectively;
those for subtask B are shown in Table \ref{tab:TaskBTwitter} for Twitter and in Table \ref{tab:TaskBSMS} for SMS messages.
Systems are ranked by their scores for the constrained runs;
the ranking based on scores for unconstrained runs is shown as a subindex.


For both subtasks, there were teams that only submitted results for the Twitter test set.
Some teams submitted both a constrained and an unconstrained version (e.g., AVAYA and teragram).
As one would expect,
the results on the Twitter test set tended to be better than those on the SMS test set
since the SMS data was out-of-domain with respect to the training (Twitter) data.

Moreover, the results for subtask A were significantly better than those for subtask B, which shows that it is a much easier task,
probably because there is less ambiguity at the phrase-level.

\begin{table}[t!]
    \centering
    \begin{small}
    \begin{tabular}{l@{ }c@{ }c@{ }c@{ }c}
    \bf Run & \bf Const- & \bf Unconst- & \bf Use & \bf Super-\\
    & \bf rained & \bf rained & \bf Neut.? & \bf vised?\\
    \hline
    NRC-Canada & 69.02 &  & yes & yes\\
    GU-MLT-LT & 65.27 &  & yes & yes\\
    teragram & 64.86 & 64.86$_{(1)}$ & yes & yes\\
    BOUNCE & 63.53 &  & yes & yes\\
    KLUE & 63.06 &  & yes & yes\\
    AMI\&ERIC & 62.55 & 61.17$_{(3)}$ & yes & yes/semi\\
    FBM & 61.17 &  & yes & yes\\
    AVAYA & 60.84 & 64.06$_{(2)}$ & yes & yes/semi\\
    SAIL & 60.14 & 61.03$_{(4)}$ & yes & yes\\
    UT-DB & 59.87 &  & yes & yes\\
    FBK-irst & 59.76 &  & yes & yes\\
    nlp.cs.aueb.gr & 58.91 &  & yes & yes\\
    $^\diamond$UNITOR & 58.27 & 59.50$_{(5)}$ & yes & semi\\
    LVIC-LIMSI & 57.14 &  & yes & yes\\
    Umigon & 56.96 &  & yes & yes\\
    NILC\_USP & 56.31 &  & yes & yes\\
    DataMining & 55.52 &  & yes & semi\\
    $^\diamond$ECNUCS & 55.05 & 58.42$_{(6)}$ & yes & yes\\
    nlp.cs.aueb.gr & 54.73 &  & yes & yes\\
    ASVUniOfLeipzig & 54.56 &  & yes & yes\\
    SZTE-NLP & 54.33 & 53.10$_{(9)}$ & yes & yes\\
    CodeX & 53.89 &  & yes & yes\\
    Oasis & 53.84 &  & yes & yes\\
    NTNU & 53.23 & 50.71$_{(10)}$ & yes & yes\\
    UoM & 51.81 & 45.07$_{(15)}$ & yes & yes\\
    SSA-UO & 50.17 &  & yes & no\\
    SenselyticTeam & 50.10 &  & yes & yes\\
    UMCC\_DLSI\_(SA) & 49.27 & 48.99$_{(12)}$ & yes & yes\\
    bwbaugh & 48.83 & 54.37$_{(8)}$ & yes & yes/semi\\
    senti.ue-en & 47.24 & 47.85$_{(13)}$ & yes & yes\\
    SU-sentilab &  & 45.75$_{(14)}$ & yes & yes\\
    OPTWIMA & 45.40 & 54.51$_{(7)}$ & yes & yes\\
    REACTION & 45.01 &  & yes & yes\\
    uottawa & 42.51 &  & yes & yes\\
    IITB & 39.80 &  & yes & yes\\
    IIRG & 34.44 &  & yes & yes\\
    sinai & 16.28 & 49.26$_{(11)}$ & yes & yes\\
       \hline
    Majority Baseline & \multicolumn{2}{c}{29.19}  & N/A & N/A \\
    \hline
    \end{tabular}
    \end{small}
    \caption{\label{tab:TaskBTwitter}Results for subtask B on the Twitter dataset.
             The $^\diamond$ indicates a system submitted as constrained but which used additional Tweets or additional sentiment-annotated text
             to collect statistics that were then used as a feature.}
  \end{table}

\begin{table}[t!]
    \centering
    \begin{small}
    \begin{tabular}{l@{ }c@{ }c@{ }c@{ }c}
    \bf Run & \bf Const- & \bf Unconst- & \bf Use & \bf Super-\\
    & \bf rained & \bf rained & \bf Neut.? & \bf vised?\\
    \hline
    NRC-Canada & 68.46 &  & yes & yes\\
    GU-MLT-LT & 62.15 &  & yes & yes\\
    KLUE & 62.03 &  & yes & yes\\
    AVAYA & 60.00 & 59.47$_{(1)}$ & yes & yes/semi\\
    teragram &  & 59.10$_{(2)}$ & yes & yes\\
    NTNU & 57.97 & 54.55$_{(6)}$ & yes & yes\\
    CodeX & 56.70 &  & yes & yes\\
    FBK-irst & 54.87 &  & yes & yes\\
    AMI\&ERIC & 53.63 & 52.62$_{(7)}$ & yes & yes/semi\\
    $^\diamond$ECNUCS & 53.21 & 54.77$_{(5)}$ & yes & yes\\
    UT-DB & 52.46 &  & yes & yes\\
    SAIL & 51.84 & 51.98$_{(8)}$ & yes & yes\\
    $^\diamond$UNITOR & 51.22 & 48.88$_{(10)}$ & yes & semi\\
    SZTE-NLP & 51.08 & 55.46$_{(3)}$ & yes & yes\\
    SenselyticTeam & 51.07 &  & yes & yes\\
    NILC\_USP & 50.12 &  & yes & yes\\
    REACTION & 50.11 &  & yes & yes\\
    SU-sentilab &  & 49.57$_{(9)}$ & no & yes\\
    nlp.cs.aueb.gr & 49.41 & 55.28$_{(4)}$ & yes & yes\\
    LVIC-LIMSI & 49.17 &  & yes & yes\\
    FBM & 47.40 &  & yes & yes\\
    ASVUniOfLeipzig & 46.50 &  & yes & yes\\
    senti.ue-en & 44.65 & 46.72$_{(12)}$ & yes & yes\\
    SSA\_UO & 44.39 &  & yes & no\\
    UMCC\_DLSI\_(SA) & 43.39 & 40.67$_{(14)}$ & yes & yes\\
    UoM & 42.22 & 35.22$_{(15)}$ & yes & yes\\
    OPTWIMA & 40.98 & 47.15$_{(11)}$ & yes & yes\\
    uottawa & 40.51 &  & yes & yes\\
    bwbaugh & 39.73 & 43.43$_{(13)}$ & yes & yes/semi\\
    IIRG & 22.16 &  & yes & yes\\
       \hline
    Majority Baseline & \multicolumn{2}{c}{19.03}  & N/A & N/A \\
    \hline
    \end{tabular}
    \end{small}
    \caption{\label{tab:TaskBSMS}Results for subtask B on the SMS dataset.
             The $^\diamond$ indicates a system submitted as constrained but which used additional Tweets or additional sentiment-annotated text
             to collect statistics that were then used as a feature.}
  \end{table}

\subsection{Subtask A: Contextual Polarity}

Table \ref{tab:TaskATwitter} shows that subtask A, Twitter,
attracted 23 teams, who submitted 21 constrained and 7 unconstrained systems.
Five teams submitted both a constrained and an unconstrained system,
and two other teams submitted constrained systems that are on the boundary between being constrained and unconstrained.

One system was semi-supervised, and the rest were supervised.
The supervised systems used classifiers such as SVM (8 systems), Naive Bayes (7 systems), and Maximum Entropy (3 systems). Other approaches used include an ensemble of classifiers, manual rules, and a linear classifier. Two of the systems chose not to predict neutral as a possible classification label.


The average F1-measure on the Twitter test set was 74.1\% for constrained systems and 60.5\% for unconstrained ones; this does not mean that using additional data does not help, it just shows that the best teams only participated with a constrained system.
NRC-Canada had the best constrained system with an F1-measure of 88.9\%, and AVAYA had the best unconstrained one with F1=87.4\%.

Table~\ref{tab:TaskASMS} shows the results for the SMS test set,
where 20 teams submitted 19 constrained and 7 unconstrained systems
(again, this included two teams that submitted boundary systems, marked accordingly).
The average F-measure on this test set was 70.8\% for constrained systems and 65.7\% for unconstrained systems. The best constrained system was that of GU-MLT-LT with an F-measure of 88.4\%, and AVAYA had the best unconstrained system with an F1 of 85.8\%.

\subsection{Subtask B: Message Polarity}

Table \ref{tab:TaskBTwitter} shows that subtask B, Twitter,
attracted 38 teams, who submitted 36 constrained and 15 unconstrained systems (and two boundary ones).

The average F1-measure was 53.7\% for the constrained and 54.6\% for the unconstrained systems.

These averages are much lower than those for subtask A,
which indicates that subtask B is harder,
probably because a message can contain parts expressing both positive and negative sentiment.

Once again, NRC-Canada had the best constrained system with an F1-measure of 69\%, followed by teragram, which had the best unconstrained system with an F1-measure of 64.9\%.

As Table \ref{tab:TaskBSMS} shows,
the average F1-measure on the SMS test set was 50.2\% for constrained and 50.3\% for unconstrained systems.
NRC-Canada had the best constrained system with an F1=68.5\%, and AVAYA had the best unconstrained one with F1-measure of 59.5\%.

\subsection{Overall}

Overall, the results achieved by the best teams were very strong, especially for the simpler subtask A:

\begin{itemize}
    \item F1=88.93, NRC-Canada on subtask A, Twitter;
    \item F1=88.37, GU-MLT-LT on subtask A, SMS;
    \item F1=69.02, NRC-Canada on subtask B, Twitter;
    \item F1=68.46, NRC-Canada on subtask B, SMS.
\end{itemize}

We can see that the strongest team overall was that of NRC-Canada, which was ranked first on three of the four conditions;
and it was second on subtask A, SMS.
There were two other teams that were strong across both tasks and on both test sets: GU-MLT-LT and AVAYA.
Three other teams, namely teragram, BOUNCE and KLUE, were ranked in the top-3 in at least one subtask and test set.

\section{Discussion}
\label{sec-discussion}


We have seen that most participants restricted themselves to the provided data and submitted constrained systems.
Indeed, the best systems for each of the two subtasks and for each of the two testing datasets were constrained systems;
of course, this does not mean that additional data would not be useful.
Curiously, in some cases where a team submitted a constrained and unconstrained run,
the unconstrained run actually performed worse.

Not surprisingly, most systems were supervised;
there were only five semi-supervised systems, and there was only one unsupervised system.
One additional team declared their system as unsupervised
since it was not making use of the training data;
we still classified it as supervised though since it did use supervision -- in the form of manual rules.

Most participants predicted all three labels (positive, negative and neutral),
even though some participants opted for not predicting neutral, which made some sense since the final F1-score
was averaged over the positive and the negative predictions only.

The most popular classifiers included SVM, MaxEnt, linear classifier, Naive Bayes;
in some cases, manual rules or ensembles of classifiers were used.

A variety of features were used, including
word-related (e.g., words, stems, $n$-grams, word clusters),
word-shape (e.g., punctuation, capitalization),
syntactic (e.g., POS tags, dependency relations),
Twitter-specific (e.g., repeated characters, emoticons, URLs, hashtags, slang, abbreviations),
and sentiment-related (e.g., negation);
one team also used discourse relations.
Almost all participants relied heavily of various sentiment lexicons,
the most popular ones being MPQA and SentiWordNet,
as well as AFINN and Bing Liu's Opinion Lexicon;
some participants used their own lexicons -- preexisting or built from the provided data.

Given that Twitter messages are noisy, most participants did some preprocessing, including
tokenization,
stemming, lemmatization,
stopword removal,
normalization/removal of URLs, hashtags, users, slang, emoticons, repeated vowels, punctuation;
some even did pronoun resolution.

\section{Conclusion}
\label{sec-conclusions}
We have described a new task that entered SemEval-2013: task 2 on Sentiment Analysis on Twitter.
The task has attracted a very high number of participants: 149 submissions from 44 teams.

We believe that the datasets that we have created as part of the task and which we have
released to the community\footnote{\texttt{http://www.cs.york.ac.uk/semeval-2013/task2/}} under a
Creative Commons Attribution 3.0 Unported License,\footnote{\texttt{http://creativecommons.org/licenses/by/3.0/}}
will be found useful by researchers beyond SemEval.

\section*{Acknowledgments}

The authors would like to thank Kathleen McKeown for her insight in creating the Amazon Mechanical Turk annotation task.


Funding for the Amazon Mechanical Turk annotations was provided by the JHU Human Language Technology Center of Excellence and by
 the Office of the Director of National Intelligence (ODNI), Intelligence Advanced Research Projects Activity (IARPA), through the U.S. Army Research Lab. All statements of fact, opinion or conclusions contained herein are those of the authors and should not be construed as representing the official views or policies of IARPA, the ODNI or the U.S. Government.

\bibliographystyle{acl}

\bibliography{naaclhlt2013}

\end{document}